%% file: acl_latex.tex
\pdfoutput=1

\documentclass[11pt]{article}

\usepackage[preprint]{acl}

\usepackage{times}
\usepackage{latexsym}

\usepackage[T1]{fontenc}

\usepackage[utf8]{inputenc}

\usepackage{microtype}

\usepackage{inconsolata}

\usepackage{graphicx}

\usepackage{booktabs}
\usepackage{amsmath}
\usepackage{graphicx}
\DeclareGraphicsExtensions{.pdf,.png,.jpg}
\usepackage{amsmath}
\usepackage{amssymb}
\usepackage{multirow}
\usepackage{pgfplotstable}
\usepgfplotslibrary{statistics}
\usepackage{pgfplots}
\pgfplotsset{width=8cm,compat=1.9}
\usepackage{bbold}
\usepackage{amsmath}
\usepackage{booktabs} 
\usepackage{algorithm}
\usepackage{algpseudocode}

\usepackage{kotex}
\title{ControlMed: Adding Reasoning Control to Medical Language Model}

\author{Sung-Min Lee \;\;\; Siyoon Lee \;\;\; Juyeon Kim \;\;\; Kyoungmin Roh \\
Agentic AI Lab, KT \\
\small\texttt{\{lee.sungmin, siyoon.lee, jy95.kim, kyoungmin.roh\}@kt.com} }

\begin{document}
\maketitle
\begin{abstract}
Reasoning Large Language Models (LLMs) with enhanced accuracy and explainability are increasingly being adopted in the medical domain, as the life-critical nature of clinical decision-making demands reliable support. Despite these advancements, existing reasoning LLMs often generate unnecessarily lengthy reasoning processes, leading to significant computational overhead and response latency. These limitations hinder their practical deployment in real-world clinical environments. To address these challenges, we introduce \textbf{ControlMed}, a medical language model that enables users to actively control the length of the reasoning process at inference time through fine-grained control markers. ControlMed is trained through a three-stage pipeline: 1) pre-training on a large-scale synthetic medical instruction dataset covering both \textit{direct} and \textit{reasoning responses}; 2) supervised fine-tuning with multi-length reasoning data and explicit length-control markers; and 3) reinforcement learning with model-based reward signals to enhance factual accuracy and response quality. Experimental results on a variety of English and Korean medical benchmarks demonstrate that our model achieves similar or better performance compared to state-of-the-art models. Furthermore, users can flexibly balance reasoning accuracy and computational efficiency by controlling the reasoning length as needed. These findings demonstrate that ControlMed is a practical and adaptable solution for clinical question answering and medical information analysis.
\end{abstract}

\begin{figure}[t!]
  \includegraphics[width=\columnwidth]{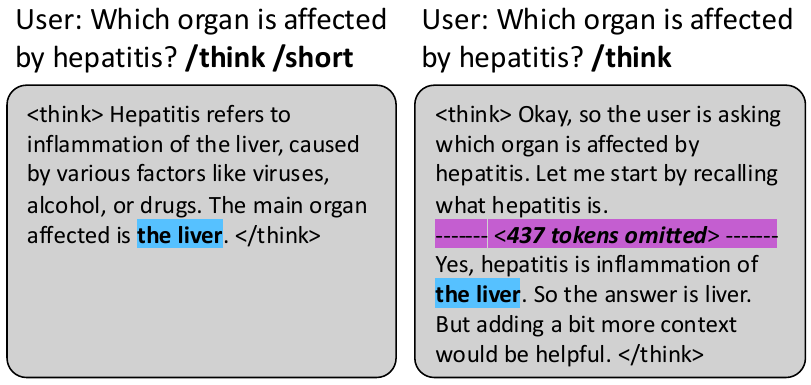}
  \caption{An illustration of the effect of different reasoning control markers on model behavior. When prompted with the '\texttt{/short}' reasoning marker, ControlMed generates more concise reasoning compared to the original \textit{reasoning mode} while maintaining answer correctness.~(\textcolor{violet}{purple}: omitted reasoning tokens, \textcolor{blue}{blue}: ground-truth answer)} 
  \label{fig:fig1_comparison_diff_reas_marker}
\end{figure}

\section{Introduction}
Large language models~(LLMs) have achieved outstanding results in a wide range of medical natural language processing tasks, such as medical question answering, clinical diagnosis, and patient consultation~\cite{su2025applications}. Their capability to synthesize complex medical knowledge and provide detailed explanations makes them promising tools in both clinical practice and medical education~\cite{li2024scoping, singhal2025toward}. Recent advances have led to the emergence of \emph{medical reasoning models}, such as HuatuoGPT-o1~\cite{chen2024huatuogpt}, which further enhance accuracy and reliability by explicitly generating a reasoning process before providing a response.

Although sufficiently long reasoning processes have been shown to improve performance~\cite{wei2022chain, snell2024scaling}, current reasoning models still face a critical limitation in that their outputs are often excessively verbose. This unnecessary verbosity increases computational costs and reduces readability. As a result, the trade-off between reasoning length and accuracy poses significant challenges. Medical professionals require quick, actionable insights, and deployment in environments with limited resources further exacerbates this issue.

To address these challenges, we introduce \textbf{ControlMed}, a medical language model with explicit and fine-grained reasoning control. ControlMed enables users to dynamically adjust the length of its reasoning at inference time, allowing them to balance response accuracy and computational efficiency (Figure \ref{fig:fig1_comparison_diff_reas_marker}). Our approach is structured around a three-stage training pipeline:
\begin{enumerate}
\item \textbf{Pre-training for medical domain specialization and hybrid reasoning:} We construct a large-scale, high-quality synthetic medical instruction dataset using a hybrid LLM-driven pipeline. This dataset contains both direct responses and explicit reasoning sequences, filtered and verified to ensure linguistic and factual consistency.
\item \textbf{Supervised fine-tuning for reasoning length control:} We create a multi-length reasoning dataset by condensing long reasoning processes into \textsc{short}, \textsc{medium}, and \textsc{long} variants using LLMs. By fine-tuning on these examples with special reasoning-length markers, ControlMed learns to generate responses with controllable reasoning length.
\item \textbf{Reinforcement learning for improving model performance:} We further enhance our model using reinforcement learning with a model-based reward function, optimizing for both factual consistency and response quality in the medical domain.
\end{enumerate}

Experimental results on medical benchmarks demonstrate that ControlMed outperforms existing medical and general LLMs, while uniquely providing adaptive reasoning length control.

The contributions of this paper are summarized as follows: 
1) To the best of our knowledge, this work is the first to introduce explicit, fine-grained reasoning length control into medical language model;  
2) We present a three-stage training pipeline for constructing a hybrid medical language model with controllable reasoning;
3) Extensive experiments on various English and Korean medical benchmarks demonstrate that our proposed ControlMed achieves superior performance while enabling practical control over reasoning length, reducing inference cost with minimal performance loss.

\begin{figure*}[ht!]

  \centering
  \includegraphics[width=0.85\linewidth]{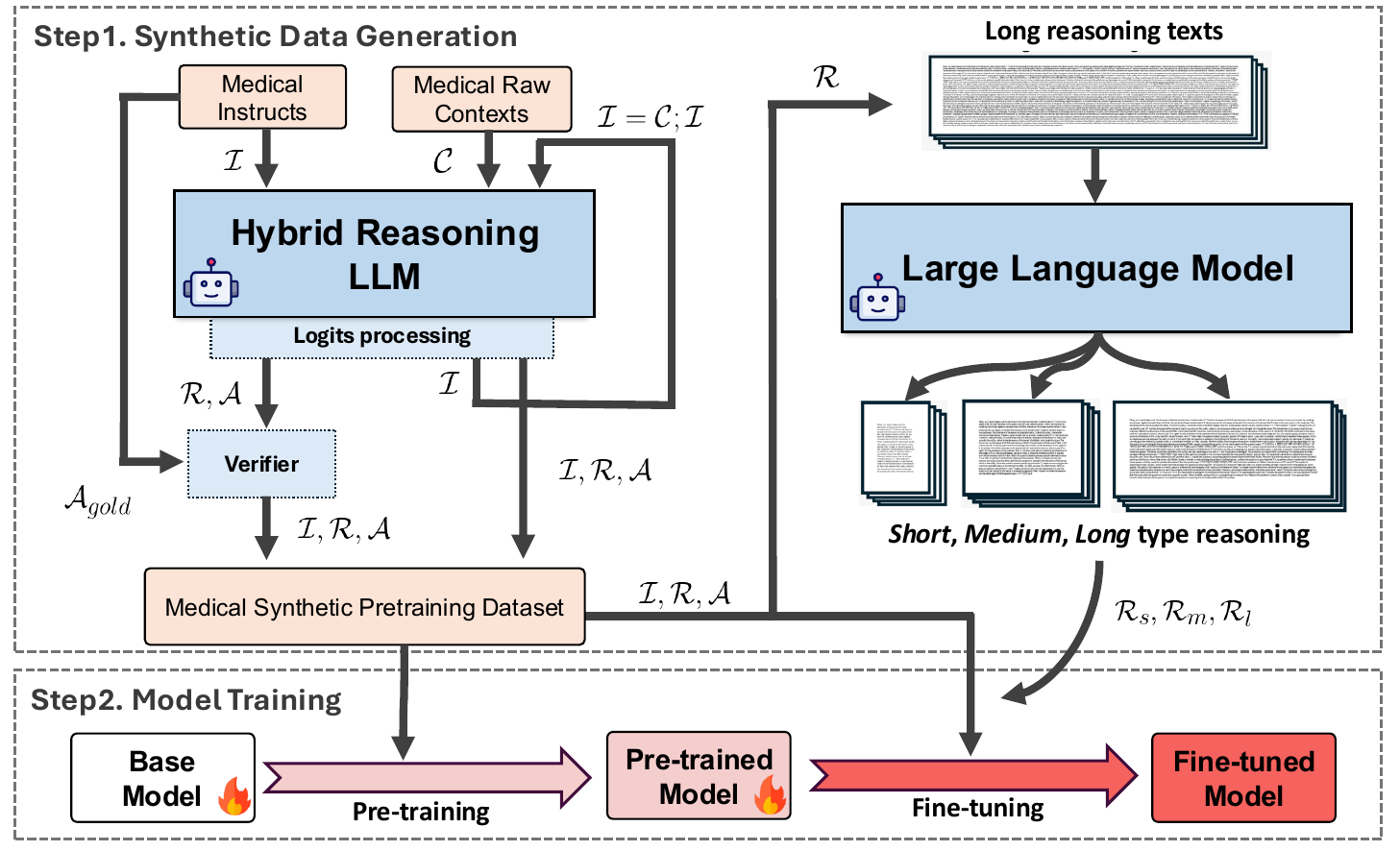}
    \caption{The overall pipeline for training ControlMed. Step 1 (Data Generation): A medical synthetic pretraining dataset is constructed using a hybrid reasoning LLM, which generates both reasoning and non-reasoning responses from medical instructions ($\mathcal{I}$) and raw medical contexts. Logit-level filtering is applied to restrict foreign language tokens. A verifier LLM further assesses the alignment between generated responses ($\mathcal{A}$) and gold responses ($A_{\mathrm{gold}}$) to ensure data quality. The resulting dataset consists of triplets $(\mathcal{I}, \mathcal{R}, \mathcal{A})$, where $\mathcal{R}$ denotes reasoning sequences. To construct the finetuning dataset for reasoning length control, long reasoning texts are condensed into short, medium, and long variants ($\mathcal{R}_s, \mathcal{R}_m, \mathcal{R}_l$) using a large language model. Step 2 (Model Training): A base model is first pre-trained using the pretraining dataset with a hybrid loss that unifies reasoning and direct response modes, and is subsequently fine-tuned on the multi-length dataset with special markers (\texttt{/short}, \texttt{/medium}, \texttt{/long}) for reasoning length control. After finetuning, we further deploy reinforcement learning (RL) to improve model performance.}
  \label{fig:architecture}
\end{figure*}

\section{Related Works}
\subsection{Medical Specialized Language Models}
Recent advancements in medical LLMs have predominantly centered on continued pre-training and instruction tuning utilizing extensive biomedical corpora and synthetic medical datasets to enhance domain-specific expertise~\cite{zhang2024ultramedical, xie2024me, peng2023study, kim2025small, sellergren2025medgemma}. As a result, medical LLMs consistently outperform their general-purpose counterparts across a range of clinical benchmarks. Also, several studies have prioritized expanding the multilingual and cross-lingual capabilities of medical LLMs~\cite{yano-etal-2025-elaine, wang2024apollo}. 

Furthermore, to address more complex and hard medical problems, medical reasoning models such as HuatuoGPT-o1~\cite{chen2024huatuogpt}, MedReason~\cite{wu2025medreason}, and CoD~\cite{chen-etal-2025-cod} have been developed to explicitly generate the reasoning process. These reasoning models not only provide greater transparency and interpretability but also achieve higher accuracy by explicitly decomposing complex clinical problems into structured reasoning processes before generating final responses.

\subsection{Length Control in Reasoning Models}
Recent research has shown that increasing the length of reasoning processes generally improves the performance of reasoning models across a variety of tasks~\cite{snell2024scaling, guo2025deepseek, yang2025qwen3}. However, this improvement often comes at the expense of increased computational cost, which can limit the practicality of deploying such models in real-world scenarios. Importantly, the effectiveness of extended reasoning is highly task-dependent: while complex tasks benefit substantially from longer reasoning chains, simpler tasks often achieve optimal performance with much shorter reasoning steps~\cite{jin-etal-2024-impact}. To address this, several approaches have been proposed to adaptively control reasoning length according to task difficulty~\cite{jie2023prompt}. Most existing methods regulate reasoning length either implicitly during training via reinforcement learning~\cite{fang2025thinkless, wan2025qwenlong}, or through decoding-time interventions~\cite{muennighoff2025s1}.

In contrast to existing work, ControlMed provides explicit and fine-grained control over reasoning length at inference time via the use of control markers, rather than relying solely on decoding-time heuristics or RL-based adaptation. Moreover, ControlMed is specifically optimized for the medical domain and supports bi-lingual scenarios.

\section{Methodology}
Our goal is to develop a hybrid medical language model with controllable reasoning length. In this section, we describe our proposed methods in detail. Figure~\ref{fig:architecture} shows the overall methods.

\subsection{Background: Hybrid Reasoning Model}
Suppose that $\mathcal{I}$ is the set of possible input instructions for a language model, and let $x \in \mathcal{I}$ denote a specific instruction to the model. To control the model’s reasoning behavior, a special reasoning marker sequence, denoted as \texttt{/think}, can be prepended to the instruction. We define two types of model behavior for a given instruction:

\textit{Reasoning Response}:
Given an input of the form $x_{r} = x;~\texttt{/think}$, the model produces an output $y_{r} = \textsc{M}(x_{r})$, which is expected to include both intermediate reasoning steps and a final response.

\textit{Direct Response}:
Given a standard input $x$, the model generates $y_{d} = \textsc{M}(x)$, which yields a direct answer without explicit reasoning.

Formally, the output under these two settings can be described as follows:
\begin{eqnarray}
\label{eq:lowEMA}
\textsc{M}(x_{r}) &=& \texttt{<think>};\; r ; \;\texttt{</think>};\; a  \nonumber\\ 
\textsc{M}(x) &=& a 
\end{eqnarray}
where $\textsc{M}(\cdot)$ denotes the inference function of a large language model, and the use of the \texttt{/think} marker governs whether the model follows an explicit reasoning process (\emph{reasoning mode}) or returns a direct response (\emph{non-reasoning mode}).

\subsection{Large Scale Synthetic Medical Instruction Dataset Construction}
\label{sec:pre_training_set_const}
In this section, we specifically describe the construction of a large-scale synthetic medical instruction dataset designed to train a hybrid reasoning model for medical instruction tasks.

\paragraph{LLM-based \emph{reasoning}, \emph{non-reasoning} synthetic data generation}
To construct a large-scale medical dataset encompassing both \emph{reasoning} and \emph{non-reasoning} instruction types, we leverage an open-source hybrid large language model (LLM) specifically optimized for both tasks. This approach allows us to distill not only the medical knowledge internalized by the model but also its complex reasoning abilities into our dataset. To ensure diversity and realism in instruction scenarios, we aggregate data from multiple sources, including publicly available medical instruction datasets, web-crawled medical QA pairs, specialized medical literature, and de-identified clinical records. Using these materials, we prompt the hybrid LLM to generate synthetic dataset, comprising instruction–reasoning–response triplets as well as instruction–response pairs. Specifically, we employ two strategies: (1) utilizing pre-existing medical instructions as prompts for the LLM to generate corresponding responses, and (2) implementing a two-stage pipeline that exploits rich medical contexts (such as patient records, clinical note excerpts, or literature abstracts). In the first stage, the LLM generates plausible, contextually relevant medical instructions conditioned on the provided context. In the second stage, the LLM produces high-quality responses to each synthetic instruction, grounded in its originating context.

\paragraph{Logit processing for restricting foreign languages} Since our goal is to construct English-Korean bilingual data, it is crucial to prevent the generation of responses containing tokens from unrelated languages such as Arabic, Russian, and others. To achieve this, we apply logit-level filtering before the softmax operation during sequence generation. Specifically, for any token $i$ that belongs to a set of prohibited languages $\mathcal{L}$, we set its logit to $-\infty$, while preserving the original logits for all other tokens:

\begin{eqnarray}
\label{eq_gated_FiD}
z'_i = 
\begin{cases}
    z_i, & \text{if } i  \notin \mathcal{L} \\
    -\infty, & \text{if } i \in \mathcal{L}   \nonumber
\end{cases} \\
P(i) = \frac{\exp(z'_i)}{\sum_{j=1}^{V} \exp(z'_j)}
\end{eqnarray}

This approach guarantees that tokens from restricted languages are assigned zero probability after softmax, thereby ensuring linguistic consistency in the synthetic dataset. The restricted set $\mathcal{L}$ comprises tokens from Arabic, Russian, and other non-target languages.

\paragraph{Model-based Verification}
To ensure the validity of synthetic responses, we introduce a verification step that leverages large language models (LLMs) as evaluators. Specifically, for cases in which synthetic responses are generated for existing instruction-response pairs, the verifier model is prompted to compare the original (gold) response with the synthetic response. This comparison is performed on a numerical scale from 0 to 10, reflecting the degree of alignment with the gold response. Only those synthetic responses that surpass a predetermined threshold are retained in the final dataset, thereby filtering out irrelevant or inaccurate responses.

\paragraph{Rule-based Post processing and Filtering}
To further improve the quality of the dataset, we deployed rule based post-processing and filtering. First, outputs that contain excessively repetitive sequences are identified and removed, as such patterns often indicate model degeneration or off-topic content. Additionally, duplicate instruction-response pairs are filtered out to prevent redundancy and to maintain diversity within the dataset. To ensure the professionalism and reliability of the medical content, any synthetic responses containing emojis are also processed by removing all emojis, as their inclusion may undermine clinical credibility. Collectively, these post-processing procedures result in a synthetic dataset that is coherent, professional for medical specialized tasks.

\subsection{Pre-training for Medical Domain Specialization and Hybrid Reasoning}
Leveraging the high-quality synthetic dataset described above, we pre-train a large language model to improve its specialization in the medical domain and to enhance its hybrid reasoning capabilities.

The learning objective unifies \emph{reasoning mode} and \emph{non-reasoning mode}. Specifically, for each training example, let $x$ denote the given instruction, $m$ denote a special reasoning marker, can be prepended to the instruction, and $y^{\text{reason}}$ and $y^{\text{direct}}$ the corresponding target outputs for each mode, with sequence lengths $T^{\text{reason}}$ and $T^{\text{direct}}$ respectively. The conditional negative log-likelihood losses are defined as:

\begin{align}
\mathcal{L}^{\text{reason}} &= -\sum_{t=1}^{T^{\text{reason}}} \log P(y_t^{\text{reason}} \mid x, m, y_{<t}^{\text{reason}})  \nonumber\\
\mathcal{L}^{\text{direct}} &= -\sum_{t=1}^{T^{\text{direct}}} \log P(y_t^{\text{direct}} \mid x, y_{<t}^{\text{direct}})
\end{align}

To blend both training modes, we introduce a hybrid objective that mixes the reasoning and direct losses for each sample, governed by an indicator variable $p_i \in \left\{0, 1\right\}$, which marks whether the sample is presented in \emph{reasoning mode} or \emph{non-reasoning mode}:

\begin{equation}
    \mathcal{L}^{\text{hybrid}} = \frac{1}{N} \sum_{i=1}^{N} \left(p_i\mathcal{L}^{\text{reason}}_i + (1-p_i)\mathcal{L}^{\text{direct}}_i \right)
\end{equation}

Here, $N$ denotes the batch size, and $\mathcal{L}^{\text{reason}}_i$ and $\mathcal{L}^{\text{direct}}_i$ correspond to each training pair's losses according to their designated response style.

Through exposure to this hybrid loss during pre-training, the model acquires proficiency not only in medical knowledge and domain-specific terminology, but also in dynamically adjusting its reasoning with reasoning marker.

\subsection{Supervised Fine-tuning for Reasoning Length Control}
To enable control over reasoning length at inference time, we perform supervised fine-tuning (SFT) using our synthetic dataset, which comprises examples with diverse reasoning lengths. In this section, we describe our methodology for constructing the SFT dataset and fine-tuning.

\paragraph{Multi-Length Reasoning Dataset Construction}
To construct a dataset suitable for supervised fine-tuning with explicit control over reasoning length, we first extract instances containing long reasoning from the pre-training dataset. To obtain shorter variants of these reasoning texts, we leverage a large language model (LLM) to remove redundant or unnecessary reasoning steps. Specifically, for each original reasoning text $\mathcal{R}$, we prompt the LLM with a prompt $p$ to generate a condensed version of the reasoning, constrained to a target word limit $n_{mode}$ corresponding to the desired length mode. This process is formalized as follows:
\begin{equation}
    \mathcal{R}_{mode} = \mathsf{LLM}\left( \mathcal{R}, p, n_{mode} \right)
\end{equation}
where $mode \in \{ \textsc{short},\; \textsc{medium},\; \textsc{long} \}$. By iteratively applying this procedure, we construct a multi-length reasoning dataset in which each instance is paired with three versions of the reasoning text, each adhering to a different length constraint.
\paragraph{Supervised Fine-tuning}
For supervised fine-tuning, we prepend a length-specific reasoning marker~(\texttt{/short}, \texttt{/medium}, or \texttt{/long}) to original reasoning marker~(\texttt{/think}) , corresponding to the desired reasoning length mode. Apart from this modification, the fine-tuning procedure mirrors the pre-training setup, utilizing the same hybrid loss formulation and optimization strategy. This approach enables the model to learn explicit associations between the reasoning marker and the target reasoning length. As a result, the model acquires the capability to dynamically adjust the length of its reasoning at inference time, conditioned on the provided marker.

\subsection{Reinforcement Learning for Improving Model Performance}
To further improve our model performance, we apply reinforcement learning, leveraging a model-based reward function.
\subsubsection{PPO Training}
Proximal Policy Optimization (PPO) is one of the most widely employed reinforcement learning algorithms for language models. By employing PPO, we can optimize the model to generate outputs that maximize expected rewards while restricting policy deviation within a clipped range, thus ensuring stable and efficient training. The PPO objective is defined as:

{\small
\begin{eqnarray}
&&L_{PPO}(\theta) = \\ \nonumber 
&&\mathbb{E}_t \left[ \min \left( r_t(\theta) \hat{A}_t, ; \mathrm{clip}(r_t(\theta), 1 - \epsilon, 1 + \epsilon) \hat{A}_t  \right) \right]
\end{eqnarray}
}

\begin{table*}[!hbt]
\small\centering
\resizebox{0.95\textwidth}{!}{ 
    \begin{tabular}{l|ccc|cc|cc|c}
    \toprule
        \multirow{3}{*}{\textit{Model}}
      & \multirow{3}{*}{\textbf{MedQA}}
      & \multirow{3}{*}{\textbf{MedMCQA}}
      & \multirow{3}{*}{\textbf{PubMedQA}}
      & \multicolumn{2}{c|}{\textbf{MMLU-Pro}}
      & \multicolumn{2}{c|}{\textbf{GPQA}}
      & \multirow{3}{*}{\textbf{Avg.}} \\ \cline{5-8}
       & & & & \textbf{Health} & \textbf{Biology} & \textbf{Genetics} & \begin{tabular}[c]{@{}c@{}}\textbf{Molecular}\\\textbf{Biology}\end{tabular} & \\
    \midrule
     BioMistral-7B   & 45.0 & 40.2  & 66.9 & 27.4 & 49.2 & 28.6 & 38.5 & 42.3  \\
     OpenBioLLM-8B & 57.7 & 54.1  & 74.1 & 38.4 & 52.4 & 43.7 & 39.6 & 51.4  \\
      UltraMedical-8B & 71.1 & 58.3  & 77.4 & 55.1 & 66.7 & 41.2 & 48.4 & 59.7  \\
    Mistral-7B-Instruct & 48.2 & 44.6  & 59.5 & 33.7 & 53.6 & 30.0 & 46.1 & 45.1  \\
    Yi-1.5-9B-Chat & 50.8 & 48.7  & 69.8 & 43.4 & 65.6 & 42.5 & 48.1 & 52.7  \\
    LLaMA-3.1-8B-Instruct & 58.7 & 56.0  & 75.2 & 52.7 & 64.6 & 33.8 & 46.8 & 55.4  \\
    GLM-4-9B-Chat& 58.9 & 49.8  & 73.5 & 45.5 & 65.4 & \textbf{53.8} & 41.6 & 55.5  \\
    Qwen2.5-7B-Instruct & 57.0 & 55.6  & 72.7 & 50.6 & \underline{70.2} & 36.2 & 49.7 & 56.0  \\
    Gemma2-9B & 61.8 & 55.9  & 63.3 & 55.1 & \textbf{74.9} & 35.0 & 57.4 & 57.6  \\
     HuatuoGPT-o1-8B & \underline{72.6} & \underline{60.4}  & \textbf{79.2} &  \underline{58.7} & 68.2 & \underline{48.8} & \textbf{59.7} & \textbf{63.9}  \\
     \midrule
    \midrule
     \textbf{ControlMed (Ours)} & \textbf{78.0} & \textbf{62.2}  & \underline{77.7} &  \textbf{61.4} & 68.6 & 40.2 & \underline{58.1} & \underline{63.8}  \\
    
    \bottomrule  
    \end{tabular}
}
\caption{Main results on various biomedical benchmarks. The highest scores are marked in \textbf{bold}, while the second-highest results are \underline{underlined} for clarity.}
\label{tab:main_results}
\end{table*}

where ($r_t(\theta) = \frac{\pi_\theta(a_t \mid s_t)}{\pi_{\theta_{\mathrm{old}}}(a_t \mid s_t)}$) is the probability ratio between the new and old policies, ($\hat{A}_t$) is the estimated advantage at timestep (t), and ($\epsilon$) is a hyperparameter that controls the range of the clipped objective.

\paragraph{Model-based reward function}
To provide a reward signal, we employ a reward model trained to evaluate the alignment between the model's response and the reference (gold) response. The reward model takes as input both the model's output and the gold reference and predicts whether the output is correct.

Let $logits_i$ denote the logits output by the reward model for the $i$th example. Applying the softmax function yields class probabilities:

\begin{equation}
p_i = \mathrm{softmax}(logits_i)
\end{equation}

Here, $p_i$ represents the probability assigned to the "correct" label for the $i$th response. The final reward is computed as follows:

\begin{equation}
\text{reward}_i =
\begin{cases}
1, & \text{if } p_i > 0.4 \text{ and valid response} \\
0, & \text{otherwise} \
\end{cases}
\end{equation}

A response is considered valid if it follows the predefined output pattern, such that when the \texttt{/think} marker is provided in the input, the response includes a corresponding reasoning process. This binary reward is then used to update the policy via PPO. By utilizing the model-based reward function, the model is trained to generate responses that match the gold references, thus improving factual consistency and task performance.

\begin{table*}[!hbt]
\small\centering
\resizebox{0.85\textwidth}{!}{ 
    \begin{tabular}{ll|ccc|cc|cc}
    \toprule
        \multirow{3}{*}{\textit{Model}}
      & \multirow{3}{*}{\textit{Mode}}
      & \multirow{3}{*}{\textbf{MedQA}}
      & \multirow{3}{*}{\textbf{MedMCQA}}
      & \multirow{3}{*}{\textbf{PubMedQA}}
      & \multicolumn{2}{c|}{\textbf{MMLU-Pro}}
      & \multicolumn{2}{c}{\textbf{GPQA}} \\ \cline{6-9}
      &  &  &  &  &
        \textbf{Health} & \textbf{Biology} &
        \textbf{Genetics} &
        \begin{tabular}[c]{@{}c@{}}\textbf{Molecular}\\\textbf{Biology}\end{tabular} \\
    \midrule
    w/ RL & \textsc{max}   & 78.004 & 62.204 & 77.7 & 61.369 & 68.619 & 40.2404 & 58.1104  \\
     & \textsc{long}   & 72.505 & 60.578 & 77.2 & 61.491 & 68.479 & 43.6138 & 51.9212  \\
     & \textsc{medium} & 71.17  & 59.861 & 77.3 & 60.391 & 70.85  & 38.5538 & 54.462   \\
     & \textsc{short}  & 71.327 & 59.693 & 77.5 & 57.823 & 67.503 & 26.9876 & 55.83    \\
     & x          & 64.493 & 59.191 & 76.7 & 53.178 & 68.34  & 37.108  & 48.8594  \\
    \midrule
    \midrule
    w/ FT & \textsc{max}   & 76.433 & 62.801 & 77 & 63.08 & 70.99 & 37.3488 & 57.1328  \\
        & \textsc{long}   & 72.191 & 59.717 & 77.3 & 59.413 & 67.782 & 37.1082 & 57.3284  \\
        & \textsc{medium} & 71.406 & 60.124 & 76.4 & 61.124 & 71.269 & 33.7342 & 58.892  \\
        & \textsc{short}  & 72.427 & 59.956 & 77.4 & 60.024 & 68.34  & 30.602 & 55.244   \\
        & x          & 64.886 & 59.072 & 76.5 & 53.545 & 67.782 & 39.2768 & 48.7292  \\
    \midrule
    \midrule
    w/ PT       & \textsc{max}   & 77.69  & 62.084 & 78.6 & 62.713 & 71.129 & 34.6984 & 58.4358  \\
           & x          & 64.179 & 59.478 & 77.8 & 54.156 & 68.758 & 41.4454 & 49.5762  \\
    \midrule
    \midrule
    Base & x         & 58.7 & 56.0  & 75.2 & 52.7 & 64.6 & 33.8 & 46.8   \\
    \bottomrule
    \end{tabular}
}
\caption{Ablation study results comparing the impact of different training strategies and inference modes across multiple biomedical benchmarks.}
\label{tab:ablation_studies}
\end{table*}

\section{Experiments}
\subsection{Experimental Setup}
The details of the implementation and experiment setup is presented in Appendix \ref{sec:imp_details}. 

\paragraph{Baselines}
We compared ControlMed with the following models:
\begin{itemize}
    \setlength\itemsep{0em}
    \item \textbf{General LLMs:} We include a set of strong general-purpose large language models, namely, Qwen-2.5~\cite{yang2025qwen3}, LLaMA-3.1~\cite{grattafiori2024llama}, Gemma 2~\cite{team2024gemma}, Yi~\cite{young2024yi}, and Mistral~\cite{Jiang2023Mistral7}, for a comprehensive comparison.
    \item \textbf{Medical specialized LLMs:} We assess several domain-adapted models, including UltraMedical~\cite{zhang2024ultramedical}, OpenBioLLM~\cite{OpenBioLLMs}, and BioMistral~\cite{labrak-etal-2024-biomistral}, each of which is pre-trained or fine-tuned on large-scale biomedical corpora to better model domain-specific knowledge.
    \item \textbf{Reasoning Medical LLMs:} HuatuoGPT-o1~\cite{chen2024huatuogpt} is a specialized model designed to enhance medical reasoning by leveraging instruction tuning and reinforcement learning.
\end{itemize}

\subsection{Main Results}
Table~\ref{tab:main_results} presents the performance of ControlMed and baseline models across a variety of English medical benchmarks, including MedQA~\cite{jin2021disease}, MedMCQA~\cite{pal2022medmcqa}, PubMedQA~\cite{jin-etal-2019-pubmedqa}, MMLU-Pro~\cite{wang2024mmlu}, and GPQA~\cite{rein2024gpqa}. Overall, ControlMed achieves strong performance on all considered tasks. Compared to general-purpose LLMs and domain-specialized LLMs, ControlMed consistently demonstrates superior accuracy, particularly on challenging medical reasoning tasks such as MedQA and MedMCQA. On MedQA, for example, ControlMed achieves an accuracy of 78.0, outperforming the next best medical LLM~(HuatuoGPT-o1, 72.6) by a margin of 5.4 points. Similarly, ControlMed records the highest or second-highest scores on PubMedQA, MMLU-Pro Health, and GPQA Molecular Biology, highlighting the robustness of our approach across diverse clinical domains. 

Notably, ControlMed demonstrates competitive performance compared to HuatuoGPT-o1~(63.9 vs. 63.8 on average), which also offers reasoning capabilities, while additionally providing enhanced features such as hybrid capabilities and explicit reasoning length control during inference.

\begin{table}[tb]
    \small
    \centering
    \resizebox{0.42\textwidth}{!}{ 
    \begin{tabular}{l|c}
        \toprule
        \textit{Model} & \textbf{Score} \\
        \midrule
        HyperCLOVAX-SEED-Text-Instruct-1.5B & 37.24 \\
        Llama-3.1-8B-Instruct & 43.31  \\        
        SOLAR-10.7B-Instruct-v1.0  & 44.61 \\        
        HuatuoGPT-o1-8B  & 54.29 \\
        EXAONE-3.5-7.8B-Instruct &  56.06\\     
        \midrule
        \midrule
        \textbf{ControlMed~(Ours)} & \textbf{57.47}  \\
        \bottomrule
    \end{tabular}}
    \caption{Evaluation results on kormedmcqa benchmark.}
    \label{tab:kor_eval}
\end{table}

\subsection{Analysis}
\paragraph{Performance on Korean Medical Benchmark}
To assess the ControlMed's bi-lingual capabilities, we evaluate ControlMed on the KorMedMCQA benchmark. As shown in Table~\ref{tab:kor_eval}, ControlMed consistently outperforms all general, medical, and korean-specialized models across across all sub-domains, achieving an average score of 57.5. This demonstrates the effectiveness of our bi-lingual data construction and training methodology, enabling the model to generalize robustly to Korean medical tasks in addition to English benchmarks. The results highlight ControlMed's potential for deployment in multilingual clinical and educational settings.

\noindent \textbf{Impact of \textit{Reasoning Mode}}
As shown in Table~\ref{tab:ablation_studies}, enabling \textit{reasoning mode}~(\textsc{max})\footnote{Here, \textsc{max} denotes original reasoning mode with base reasoning marker~(\texttt{/think})} consistently yields remarkable improvements over the \textit{non-reasoning} counterparts. Specifically, \textit{reasoning mode} achieves 78.0 on MedQA and 62.2 on MedMCQA, compared to 64.5 and 59.2, respectively, for the \textit{non-reasoning mode}. These results highlight the importance of explicit reasoning supervision in promoting robust medical understanding and complex problem-solving capabilities.

\begin{figure}[tb]
    \centering
  \includegraphics[width=0.9\columnwidth]{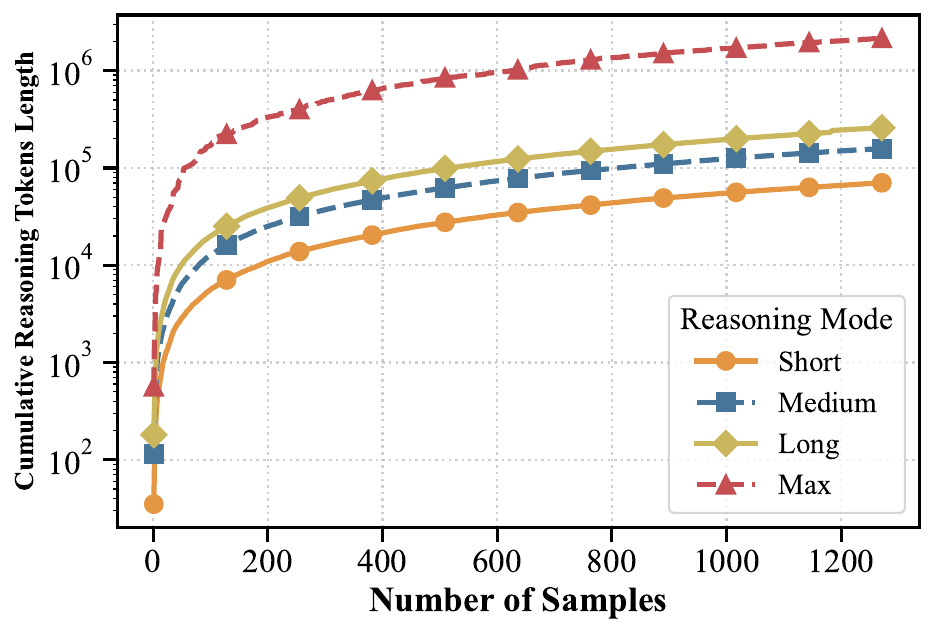}
  \caption{An illustration of the cumulative reasoning token length across different reasoning control modes (\textsc{short}, \textsc{medium}, \textsc{long}, \textsc{max}) as the number of samples increases.} 
  \label{tab:cumulative_reasoning_token_length}
\end{figure}

\noindent \textbf{Impact of Training Methods}
Pre-training with our large-scale synthetic medical dataset yields substantial performance improvements over the base model across all benchmarks, confirming the effectiveness of domain adaptation. While supervised fine-tuning with multi-length reasoning data brings only marginal additional gains in accuracy, it equips the model with explicit reasoning length control capabilities. Notably, applying reinforcement learning further boosts performance, especially on challenging tasks, increasing MedQA accuracy from 76.4 to 78.0 and GPQA Genetics from 37.3 to 40.2. These results demonstrate that each training stage contributes uniquely to both the accuracy and controllability of ControlMed.

\begin{figure}[t!]
    \begin{minipage}[c]{0.1\linewidth}
        \begin{tikzpicture}[scale=0.45]
        \centering
        \begin{axis}[
            xlabel={Reasoning Mode},
            ylabel={Score},
            symbolic x coords={x, Short, Long, Max}, 
            ymin=47, ymax=87,
            ytick={50,60,70,80},     
            xtick=data,
            legend pos=north west,
            ymajorgrids=true,
            grid style=dashed,
            scaled ticks=false, 
            tick label style={/pgf/number format/fixed},
            y tick label style={font=\footnotesize},
            x tick label style={font=\footnotesize, text width=1.1cm, align=center},
            every axis plot/.append style={thick},
        ]
        
        \addplot[
            color=blue,
            mark=*,
            ]
            coordinates {
            (x,64.49)(Short,71.32)(Long,72.50)(Max,78.00)
            };
        \addplot[
            color=purple,
            mark=x,
            ]
            coordinates {
            (x,59.19)(Short,59.69)(Long,60.57)(Max,62.20)
            };        
        \addplot[
            color=orange, %
            mark=diamond*,
            ]
            coordinates {
            (x,53.17)(Short,57.82)(Long,61.49)(Max,61.36)
            };

        \legend{MedQA, MedMCQA, MMLU-Pro Health}
        
        \end{axis}
        \end{tikzpicture}
    \end{minipage}
\hfill
    \begin{minipage}[c]{0.5\linewidth}
        \begin{tikzpicture}[scale=0.45]
        \centering
        \begin{axis}[
            xlabel={Reasoning Mode},
            ylabel={Time To First Token (s)},
            symbolic x coords={x, Short, Long, Max}, 
            ymin=0, ymax=13,
            ytick={3,6,9,12},    
            xtick=data,
            legend pos=north west,
            ymajorgrids=true,
            grid style=dashed,
            scaled ticks=false, 
            tick label style={/pgf/number format/fixed},
            y tick label style={font=\footnotesize},
            x tick label style={font=\footnotesize, text width=1.1cm, align=center},
            every axis plot/.append style={thick},
        ]
        
        \addplot[
            color=black,
            mark=*,
            ]
            coordinates {
            (x,1.824)(Short,4.639)(Long,5.706)(Max,9.866)
            };

        
        \end{axis}
        \end{tikzpicture}
    \end{minipage}
\caption{\label{fig:mode_effectiveness} Performance scores on medical benchmarks, and Time to First Token (TTFT) measured across different reasoning modes.
}
\end{figure}

\begin{figure}[t]
    \centering
  \includegraphics[width=0.85\columnwidth]{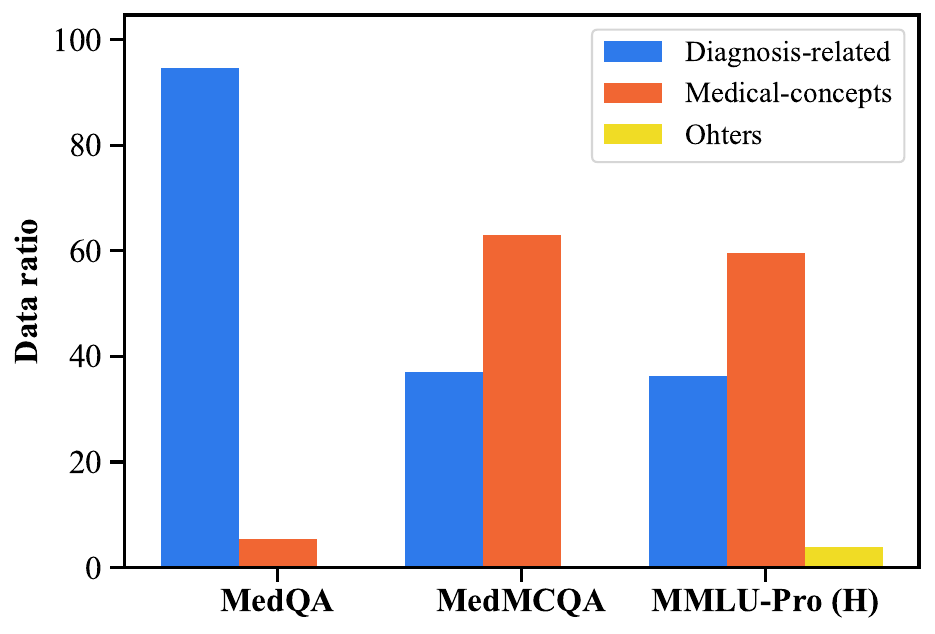}
  \caption{Statistics of medical benchmark test datasets categorized by problem type.}
  \label{fig:stats_probtype_on_benchmarks}
\end{figure}

\paragraph{Reasoning Length Control and Trade-offs}
Figure~\ref{tab:cumulative_reasoning_token_length} presents the cumulative reasoning token length across different reasoning control modes (\textsc{short}, \textsc{medium}, \textsc{long}, and \textsc{max}) as the number of samples increases. The results clearly demonstrate that ControlMed is able to control reasoning length with control marker. Specifically, the \textsc{short} mode consistently yields the lowest cumulative token count, followed by \textsc{medium} and \textsc{long}, with the \textsc{max} reasoning mode producing the most reasoning outputs. The gap between each mode remains substantial as the number of samples increases. Notably, the difference between the \textsc{short} and \textsc{max} modes spans several orders of magnitude, highlighting the practical impact of this controllability for applications with strict token or latency constraints.

This fine-grained control over reasoning length enables users to flexibly balance accuracy and computational efficiency according to specific task requirements. For example, in scenarios where rapid responses or limited computational resources are critical, the \textsc{short} mode can be employed to minimize token usage. Conversely, when more complex reasoning is necessary to ensure answer correctness, the \textsc{max} mode can be utilized. Collectively, these results validate capabilities of ControlMed to dynamically adapt its output length, thereby accommodating a wide range of downstream application needs.

\begin{table}[t!]
\centering
\small
\resizebox{0.47\textwidth}{!}{\begin{tabular}{l|r|r|r|r}
\toprule
\textbf{Department} & \textbf{\textsc{Short}} & \textbf{\textsc{Max}} & \textbf{Diff} & \textbf{Diff~(\%)} \\
\midrule
Psychiatry                  & 48  & 57  & +9  & +18.7 \\
Surgery                     & 27  & 32  & +5  & +18.5 \\
Emergency Medicine          & 72  & 77  & +5  & +6.9  \\
Internal Medicine           & 504 & 538 & +34 & +6.7  \\
Pediatrics                  & 142 & 150 & +8  & +5.6  \\
OB/GYN                      & 82  & 85  & +3  & +3.6  \\
Dermatology                 & 22  & 21  & -1  & -4.5 \\
\midrule
Otorhinolaryngology         & 1   & 2   & +1  & +100.0 \\
Ophthalmology               & 9   & 12  & +3  & +33.3 \\
Pharmacy                    & 6   & 7   & +1  & +16.6 \\

\bottomrule
\end{tabular}}

\caption{\label{fig:cccccc}Differences in the number of correct answers by department on the MedQA test dataset when using \textsc{Short} and \textsc{Max} reasoning modes, including absolute and percentage changes.}
\end{table}

\paragraph{Is Reasoning Essential to Solve Medical Problems?}
As shown in Figure~\ref{fig:mode_effectiveness}, performance on the MedQA benchmark improves noticeably as the reasoning length increases. However, for other benchmarks, the presence or length of reasoning has little impact on model performance. therefore, we conducted a more detailed analysis of the types of medical problems in each benchmark. As shown in Figure~\ref{fig:stats_probtype_on_benchmarks}, we found that the MedQA benchmark mainly consists of problems that require predicting a diagnosis based on a patient’s condition, rather than simply recalling basic medical concepts. This suggests that, unlike problems testing direct medical knowledge, tasks involving medical diagnosis demand strong reasoning capabilities.

Additionally, to better understand the effect of reasoning length, we analyzed the results by medical department within the MedQA dataset. As shown in Table~\ref{fig:cccccc}~\footnote{Departments with approximately 10 correct answers (Otorhinolaryngology, Ophthalmology, Pharmacy) are indicated separately, as differences in such small samples are not considered to be statistically meaningful.} we observe the most substantial score increases in Psychiatry and Surgery departments when longer reasoning is applied, whereas Dermatology department shows a slight decrease in performance. These observations indicate that extended reasoning length benefits solving problems in Psychiatry and Surgery departments, leading to more accurate answers, while it may not be as beneficial and can even be detrimental for some departments such as Dermatology.

\begin{table*}[t!]
\small
\centering
\resizebox{0.95\textwidth}{!}{
\begin{tabular}{p{0.1\linewidth} | p{0.9\linewidth}}
\toprule
\multicolumn{2}{c}{
\begin{tabular}[c]{@{}p{1.0\linewidth}@{}}
\textbf{\textit{Instruction}}: Please answer the following multiple-choice question: In order to assess the feasibility and evaluate the outcomes of cerclage wiring as a supportive approach to osteosynthesis in femur fractures, a group of orthopedic surgeons studied 14 patients with primary and periprosthetic fractures of the thigh bone. Parameters such as patient demographic, type of implant, number of wires used, fracture union rate, and potential complications were thoroughly recorded and analyzed in all the patients, with a mean duration of patient follow-up of 16 months. Union was achieved in all patients with a mean duration of 90 days, and there were no complications found in patients included in the study. The authors were satisfied with their findings and, due to the prospective nature of their research, submitted their study to a journal as a cohort study (which they noted in the study title as well). However, the journal editor returned the article, suggesting that it should be submitted as a case series instead. The editor made this suggestion to the authors for which of the following reasons?$\backslash$n$\backslash$nA. No prevalence assessment$\backslash$nB. Low number of patients$\backslash$nC. Ascertainment bias$\backslash$nD. Lack of risk calculation
\end{tabular}
} \\ 
\midrule
\midrule
\textbf{\textit{Mode}} & \textbf{\textit{Response}} \\
\midrule
\textsc{Medium} & 
\begin{tabular}[c]{@{}p{1.0\linewidth}@{}}
<think> A cohort study is defined by following a group over time to assess outcomes, while a case series is a descriptive report of multiple cases without comparison or control. The study described here is a prospective case series, not a cohort study, because it lacks a comparison group and does not assess risk factors or outcomes in relation to them. The editor's suggestion to change the study type to a case series is due to the lack of comparison, which is a key feature of a cohort study. Therefore, the correct answer is D: lack of risk calculation. </think> \\
 The scenario describes a prospective study involving 14 patients with femur fractures who underwent cerclage wiring. The study recorded various parameters (e.g., demographics, implant type, number of wires, union rate, complications) and followed patients for an average of 16 months.   \textbf{<511 tokens omitted>}       \  \  \   Final Answer: $\backslash$n$\backslash$n The editor suggested the study be reclassified as a case series because it lacks the essential features of a cohort study, particularly the absence of risk factor analysis or comparison groups. This aligns with the definition of a case series, which is a descriptive study without a comparative or analytical framework.$\backslash$n$\backslash$n\colorbox{cyan}{\textbf{$\backslash$boxed\{D\}}}
 \end{tabular}
   \\ \midrule 
\textsc{Max} & 
\begin{tabular}[c]{@{}p{1.0\linewidth}@{}}
<think> Okay, let's try to figure out this question. \textbf{<452 tokens omitted>} The editor's suggestion to change to a case series might be because the study is just describing the outcomes in a group without comparing to a control group or using a cohort design that follows over time. \textcolor{orange}{\textbf{\textit{Wait, but the study was prospective with a mean follow-up of 16 months. So it's a prospective cohort study. But maybe the editor thinks that with only 14 patients, it's not a proper cohort study but a case series. Cohort studies usually require a larger sample size to establish statistical significance, but 14 is small.}}} However, the key difference between a case series and a cohort is whether it's descriptive or comparative. Since the study is just describing the outcomes in 14 patients without a control group, maybe the editor considers it a case series. So the reason is the low number of patients (B) leading to it being a case series instead of a cohort. \textbf{<297 tokens omitted>} The options are A-D. So the answer is B. </think>
\\ The scenario presents a study on the use of cerclage wiring for femur fractures, involving 14 patients followed for 16 months.  \textbf{<498 tokens omitted>} \  \  \   Final Answer$\backslash$n$\backslash$n\colorbox{red}{\textbf{$\backslash$boxed\{B\}}}
\end{tabular}
\\ \bottomrule
\end{tabular}

}

\caption{A case study of ControlMed responses under different reasoning length control modes. The \textcolor{orange}{orange} highlight indicates where incorrect or misleading reasoning begins, the \textcolor{red}{red} box marks the incorrect answer, and the \textcolor{cyan}{blue} box marks the correct answer.}
\label{tab:case_study}
\end{table*}

\subsection{Case Study}
Table~\ref{tab:case_study} illustrates ControlMed’s responses to the same medical question under different reasoning length settings. In \textsc{Medium} mode, the model provides a concise and focused explanation, correctly identifying the absence of risk calculation as the key issue and thus selects the correct answer. In contrast, the \textsc{Max} mode response is significantly longer but reveals reasoning drift, as the model incorrectly fixates on the small sample size and ultimately chooses a wrong answer. This example demonstrates that long reasoning does not always guarantee higher answer accuracy and may lead the model to overthink or introduce errors.

\section{Conclusion}
In this paper, we introduced ControlMed, a novel medical language model that enables explicit and fine-grained control over the length of its reasoning process via inference-time control markers. Leveraging a three-stage training pipeline, ControlMed achieves state-of-the-art or competitive results across diverse English and Korean medical benchmarks. Our analysis demonstrates that users can flexibly trade off reasoning depth and computational cost according to the needs of specific scenarios. Furthermore, ControlMed’s bilingual capabilities highlight its applicability in multilingual clinical settings. We believe that ControlMed can serve as a valuable foundation for building reliable and adaptable medical models in real-world clinical and educational environments.

\section*{Limitations}
While ControlMed demonstrates strong performance on standard medical benchmarks, its effectiveness and safety in real-world clinical environments remain untested. To date, we have not conducted prospective deployments or rigorous user studies involving healthcare professionals and patients. Furthermore, our evaluation strategy has relied primarily on automatic metrics and publicly available datasets, which may not fully capture the clinical relevance, factual correctness, or practical utility of model outputs. Comprehensive human evaluation by domain experts is essential to assess subtle errors, reasoning quality, and the overall trustworthiness of the model in medical decision-making scenarios. To ensure safe and reliable adoption in real-world settings, future work should include structured human evaluations and prospective validation with clinicians, as well as the development of additional safeguards to mitigate risks associated with incorrect or misleading outputs.

\bibliography{anthology,custom}

\appendix

\section{Implementation Details}
\label{sec:imp_details}

\begin{table}[tb]
    \small
    \centering
    \begin{tabular}{lc}
        \toprule
        \textbf{Category} & \textbf{Number}  \\ 
        \midrule 
        \midrule
        Contextual medical instruction  &  472,683  \\
        Contextual medical instruction ($\mathcal{R}$) &  225,746  \\
        Medical instruction  &  484,310 \\
        Medical instruction ($\mathcal{R}$) &  216,845  \\
        \midrule
        Pretraining Dataset  & 1,399,584 \\
        \bottomrule
    \end{tabular}
    \caption{Statistics of the constructed synthetic pretraining dataset. $\mathcal{R}$ denotes that reasoning steps are included in the dataset.}
    \label{tab:dataset_statistics}
\end{table}

\paragraph{Datasets}
To construct the synthetic medical pretraining dataset, medical contexts are sourced from PubMed\footnote{https://pubmed.ncbi.nlm.nih.gov}, PMC-Patients case reports, and de-identified clinical narratives~\cite{zhao2022pmc}. In addition, we incorporate a diverse set of publicly available medical instruction datasets, including USMLE, ChatDoctor~\cite{li2023chatdoctor}, MedExpQA~\cite{alonso2024medexpqa}, CoD~\cite{chen2024cod}, PubMedQA~\cite{jin-etal-2019-pubmedqa}, MedQA~\cite{jin2021disease}, MedMCQA~\cite{pal2022medmcqa}, and KorMedMCQA~\cite{kweon2024kormedmcqa}, as well as web-crawled medical QA pairs. Korean medical data are included to support bilingual training for both English and Korean. Leveraging the methodology described in Section~\ref{sec:pre_training_set_const}, we synthesize a large-scale instruction–response dataset comprising both reasoning and non-reasoning samples. To generate synthetic reasoning data, we used Qwen/QwQ-32b~\footnote{https://huggingface.co/Qwen/QwQ-32B} and for reasoning text compression, we deployed OpenAI's gpt-4o. Detailed statistics of the constructed dataset are provided in Table~\ref{tab:dataset_statistics}.

For reasoning length control, we construct a multi-length reasoning dataset by extracting instances with long reasoning texts from the pretraining corpus and generating condensed variants using a large language model. Specifically, each original reasoning sequence is rewritten to produce three versions by constraining the number of words to $n_{mode}$, set to 50 for short, 150 for medium, and 700 for long, respectively.

\paragraph{Training}
As a base model of ControlMed, we used LLaMA-3.1-8b-Instruct~\footnote{https://huggingface.com/meta-llama/Llama-3.1-8B-Instruct}. To improve training efficiency, we utilize the bfloat16 mixed precision format. During pre-training, the model is trained for three epochs with a learning rate of 1e-5 and a batch size of 320. Fine-tuning is conducted for 100 steps with a learning rate of 5e-5 and a batch size of 640. Reinforcement learning is performed for one epoch, employing PPO with a learning rate of 5e-7 and a batch size of 8. All experiments are conducted using 8 NVIDIA H100 GPUs.
\paragraph{Evaluation}
For all other benchmarks, a single evaluation was performed. Due to the limited number of questions in the GPQA dataset, we conducted five separate evaluation runs on GPQA and reported the average results.

\input{table/prompt_sft_data_gen}
\input{table/promt_verification}

\begin{table*}[ht]
\small
\centering
\scalebox{0.9}{\begin{tabular}{p{0.1\linewidth} | p{0.9\linewidth}}
\toprule
\multicolumn{2}{c}{
\begin{tabular}[c]{@{}p{1.0\linewidth}@{}}
\textbf{\textit{Instruction}}: 매맞는아이증후군(어린이 구타증, battered child syndrome)에 관한 설명으로 옳은 것은?
A: 8세 이하의 여자아이에게 호발한다.
B: 부모의 진술과 임상소견이 일치한다.
C: 신체 전 부위에 장기간에 걸친 타박상을 보인다.
D: 격렬하게 치아를 꽉 깨물어 구치부 치관파절이 발생된다.
E: 구강 내 손상의 특징은 하순 소대 주위의 점막열창과 치은파열이다.
\end{tabular}
} \\ 
\midrule
\midrule
\textbf{\textit{Mode}} & \textbf{\textit{Response}} \\
\midrule
\textsc{Short} & 
\begin{tabular}[c]{@{}p{1.0\linewidth}@{}}
<think> B 옵션의 부모 진술과 임상 소견 일치는 현실적으로 불가능하므로, C(장기간 타박상)가 정답이다. </think>  **정답: C**  **해설:**  - **A. 8세 이하의 여자아이에게 호발한다.**     매맞는아이증후군은 성별과 무관하며, 남녀 모두 발생할 수 있습니다. 따라서 **틀립니다**.  - **B. 부모의 진술과 임상소견이 일치한다.**     부모의 진술은 자주 부정확하거나 진실을 왜곡할 수 있으며, 특히 학대 의심 시 진술을 부정하는 경우가 많습니다. 따라서 **틀립니다**.  - **C. 신체 전 부위에 장기간에 걸친 타박상을 보인다.**     매맞는아이증후군은 반복적인 신체적 학대로 인해 타박상, 골절, 내부 손상 등이 **장기간에 걸쳐 여러 부위에 분포**되는 것이 특징입니다. 따라서 **정답입니다**.  - **D. 격렬하게 치아를 꽉 깨물어 구치부 치관파절이 발생된다.**     치관파절은 구토 시 발생하며, 구토와의 연관성을 강조하지만 이는 매맞는아이증후군보다는 구토증후군과 관련성이 높습니다. **틀립니다**.  - **E. 구강 내 손상의 특징은 하순 소대 주위의 점막열창과 치은파열이다.**     구강 내 손상은 구토 시 발생하며, 구토와의 연관성을 강조하는데, 이는 매맞는아이증후군보다는 구토증후군과 관련성이 높습니다. **틀립니다**.  **결론:**   장기간에 걸친 전신 타박상은 매맞는아이증후군의 핵심 임상 특징이므로, **C가 정답**입니다.
 \end{tabular}\\
\midrule
\textsc{Medium} & 
\begin{tabular}[c]{@{}p{1.0\linewidth}@{}}
<think> 매맞는아이증후군은 자주 반복된 신체적 폭행으로 신체 전 부위에 타박상이 나타나는 것이 핵심 특징입니다. 이 중 C는 이 설명에 부합합니다. 각 선택지의 내용을 검토해 보면, D는 신경성 공격에 의한 전형적 손상이 아니며 E는 구강 내 손상의 일반적 특징이 아닙니다. B는 진술과 임상소견의 일치는 해당 증후군과 반드시 관련이 없으며, A는 성별 분포와도 일치하지 않습니다. 따라서 정답은 C입니다. </think>     정답: C. 신체 전 부위에 장기간에 걸친 타박상을 보인다.         설명: - A. 8세 이하의 여자아이에게 호발한다.     매맞는아이증후군은 남녀 차이 없이 발생하며, 주로 1~4세 어린이가 많지만 8세 이하 여아라는 구체적인 연령 또는 성별 선호는 없습니다. 따라서 틀립니다.  - B. 부모의 진술과 임상소견이 일치한다.     증후군 진단 시 부모의 진술이 의심스러운 경우가 많고, 임상소견과 일치하지 않을 수 있습니다. 따라서 틀립니다.  - C. 신체 전 부위에 장기간에 걸친 타박상을 보인다.     매맞는아이증후군의 핵심 특징은 반복적 신체적 폭행으로 인한 타박상, 골절, 내장 손상 등 전신 손상입니다. "장기간"이라는 표현은 반복적 폭행의 과정을 반영하며, 이는 올바른 설명입니다.  - D. 격렬하게 치아를 꽉 깨물어 구치부 치관파절이 발생된다.     "치관파절"은 외상(예: 치아 충격)으로 발생하며, 폭행과 직접적 연관이 없습니다. 따라서 틀립니다.  - E. 구강 내 손상의 특징은 하순 소대 주위의 점막열창과 치은파열이다.     구강 손상은 일반적으로 혀나 입술의 상처가 더 흔하며, 하순의 점막열창은 매맞는아이증후군의 특징이 아닙니다. 따라서 틀립니다.         결론:   C는 매맞는아이증후군의 주요 임상적 특징을 정확히 반영합니다.
 \end{tabular}
   \\ \midrule 
\textsc{Max} & 
\begin{tabular}[c]{@{}p{1.0\linewidth}@{}}
<think> 매맞는아이증후군은 부정적인 신체적 학대로 인해 발생하는 질환으로, 주로 1세에서 5세 사이의 어린이에서 흔하며, 특히 1세에서 3세 사이의 남자아이가 가장 많이 해당된다. 문제에서 제시된 보기 중, '8세 이하의 여자아이에게 호발한다(A)'는 사실과 일치하지 않으며, '부모의 진술과 임상소견이 일치한다(B)'는 상식적으로 틀린 표현이다. '격렬하게 치아를 꽉 깨물어 구치부 치관파절이 발생된다(D)'는 발음 과정 중의 상해로 인한 것이므로 매맞는아이증후군과는 관련이 없다. '구강 내 손상의 특징은 하순 소대 주위의 점막열창과 치은파열이다(E)'는 구타와는 상관이 없어 배제된다. '신체 전 부위에 장기간에 걸친 타박상을 보인다(C)'는 매맞는아이증후군의 대표적 임상 소견과 일치하므로, 정답은 'C'다. </think>     매맞는아이증후군(어린이 구타증)은 반복적인 신체적 학대로 인해 발생하는 질환으로, 주요 임상적 특징은 다음과 같습니다:         **정답: C** **C. 신체 전 부위에 장기간에 걸친 타박상을 보인다.**         **해설** 1. **A. 8세 이하의 여자아이에게 호발한다.**      - 매맞는아이증후군은 **1세에서 3세 남자아이**에서 가장 흔합니다. 따라서 여성아이가 주된 대상은 아니며, 이는 **틀립니다**.  2. **B. 부모의 진술과 임상소견이 일치한다.**      - 부모의 진술은 종종 학대와 관련이 있지만, 임상 소견(골절, 타박상 등)과 일치하지 않을 수 있습니다. 오히려 **일치하지 않는 경우가 많아** 이는 **틀립니다**.  3. **C. 신체 전 부위에 장기간에 걸친 타박상을 보인다.**      - 반복적인 학대로 인해 **다양한 부위(골반, 뇌, 뼈 등)에 만성적 손상**이 관찰됩니다. 이는 매맞는아이증후군의 주요 임상적 특징이며, **정답입니다**.  4. **D. 격렬하게 치아를 꽉 깨물어 구치부 치관파절이 발생된다.**      - 이는 **발음 과정 중의 상해**로, 학대와 직접적인 연관이 없습니다. **틀립니다**.  5. **E. 구강 내 손상의 특징은 하순 소대 주위의 점막열창과 치은파열이다.**      - 구강 내 손상은 학대와 관련될 수 있지만, **이 특정 증상은 반드시 매맞는아이증후군의 필수 증상은 아닙니다**. **틀립니다**.         **결론** 매맞는아이증후군은 **반복적 학대로 인한 다발성 타박상과 골절**이 특징입니다. 따라서 **C**가 정답입니다.
\end{tabular}
\\ \bottomrule
\end{tabular}}
\caption{A case study of ControlMed’s reasoning under different reasoning length control modes (\textsc{short}, \textsc{medium}, \textsc{max}) for a Korean clinical multiple-choice question about battered child syndrome. Across all modes, the model correctly identifies option C as the answer.}
\label{tab:RADCoT_query2doc_prompt}
\end{table*}

\end{document}

%% file: table/prompt_sft_data_gen.tex
\begin{table*}[ht]
\centering
\scalebox{0.9}{\begin{tabular}{p{0.2\linewidth} | p{0.8\linewidth}}
\toprule
Prompt & 
\begin{tabular}[c]{@{}p{1.0\linewidth}@{}}
\# Reasoning\\
$\mathsf{\{Reasoning\}}$\\
-----\\
The content above (Reasoning) describes the process by which an LLM thinks about a user's input. Please remove unnecessary parts from this reasoning process and reduce the length to within {$\mathsf{\{N_{mode}\}}$} words. Note that this is not a summary; instead, you are to streamline the thinking process by eliminating unnecessary steps, remembering that this is about a "thinking process." That is, you should sequentially contemplate the problem and state the conclusion in the final sentence. When reducing the reasoning process, you must prioritize the required elements as follows:
\\
- Priority of Elements to Include (in order of importance)\\
1. The conclusion of the reasoning\\
2. Information essential for reaching the conclusion\\
3. Additional information useful for reaching the conclusion\\
4. Useful information that is not necessary for reaching the conclusion\\
\\
Please print it out according to the Json format below. (exactly)\\
\{\\
"Reasoning" : <string>,\\
\}\\
\end{tabular}
\\ \bottomrule
\end{tabular}}
\caption{Prompt template used for generating multi-length dataset.}
\label{tab:prompt_for_sftdata}
\end{table*}

%% file: table/promt_verification.tex
\begin{table*}[ht]
\centering
\resizebox{0.95\textwidth}{!}{
\begin{tabular}{p{0.2\linewidth} | p{0.8\linewidth}}
\toprule
Prompt & 
\begin{tabular}[c]{@{}p{1.0\linewidth}@{}}
Based on the given instruction, preferred answer, and response, evaluate how well the response matches the preferred answer on a scale from 0 to 10. If the response completely deviates from or contradicts the content of the preferred answer, give a low score. You must output only one integer (0–10).\\
instruction: $\mathsf{\{Instruction\}}$\\
preferred answer: $\mathsf{\{PreferredAnswer\}}$ \\
response: $\mathsf{\{Response\}}$\\
\end{tabular}
\\ \bottomrule
\end{tabular}}

\caption{Prompt template used for LLM-based verifier.}
\label{tab:prompt_for_verification}
\end{table*}